\documentclass[sn-mathphys, iicol]{sn-jnl}

\usepackage[utf8]{inputenc}
\usepackage{xcolor}
\usepackage{amssymb}
\usepackage{float}
\usepackage{amsmath}
\usepackage{hyperref}
\usepackage{graphicx}
\usepackage{caption}
\usepackage{subcaption}
\usepackage{booktabs} 
\usepackage{algorithm,algpseudocode}
\usepackage{array}

\begin{document}

\title{Procedural terrain generation with style transfer}

\author[1]{\fnm{Fabio} \sur{Merizzi}}\email{fabio.merizzi@unibo.it}


\affil[1]{\orgdiv{Department of Informatics: Science and Engineering (DISI)}, \orgname{University of Bologna}, \orgaddress{\street{Mura Anteo Zamboni 7}, \city{Bologna}, \postcode{40126},
\country{Italy}
}}



\abstract{
In this study we introduce a new technique for the generation of terrain maps, exploiting a combination of procedural generation and Neural Style Transfer. We consider our approach to be a viable alternative to competing generative models, with our technique achieving greater versatility, lower hardware requirements and greater integration in the creative process of designers and developers. 
Our method involves generating procedural noise maps using either multi-layered smoothed Gaussian noise or the Perlin algorithm. We then employ an enhanced Neural Style transfer technique, drawing style from real-world height maps. This fusion of algorithmic generation and neural processing holds the potential to produce terrains that are not only diverse but also closely aligned with the morphological characteristics of real-world landscapes, with our process yielding consistent terrain structures with low computational cost and offering the capability to create customized maps. Numerical evaluations further validate our model's enhanced ability to accurately replicate terrain morphology, surpassing traditional procedural methods.
}

\maketitle
\noindent \textbf{keywords:} style transfer, procedural noise maps, terrain generation, morphological characteristics

\section{Introduction}

In the fields of computational graphics and geospatial simulation, terrain representation plays a crucial role in ensuring realism and accuracy. Historically, terrains have been modeled using manual design or by interpolating real-world geospatial datasets. These traditional methodologies, while effective, present challenges in scalability, adaptability, and resource intensity \cite{smelik2009survey}. The paradigm of procedurally generated terrain maps offers a solution, using algorithmic methodologies to autonomously produce terrain topographies \cite{7590336}.

Central to the process of procedural generation are sophisticated algorithms like Perlin and Simplex noise\cite{perlin2002improving} . These are particularly favored for their ability to simulate the randomness and intricacy characteristic of natural terrains. Besides offering a diverse spectrum of terrains, these algorithms also contribute to a reduction in storage requirements, as they replace voluminous terrain data with compact generative rules.

In recent years many publications showed effort in combining procedural generation techniques with more informed methods, such as software agents \cite{5454273}, erosion based approaches \cite{10.1145/2461912.2461996,mei2007fast} and genetic algorithms \cite{10.1145/1068009.1068241}. 

Recent strides in machine learning, particularly the advent of Generative Adversarial Networks (GANs) \cite{10.1145/3422622}, have ushered in a new era of potential for terrain modeling \cite{indianguy,huang2023styleterrain,beckham2017step,panagiotou2020procedural,HUANG2023373}. GANs, which pit generative models against discriminative ones in an iterative training process, have shown significant promise in refining and enhancing procedurally generated terrains. 

In this paper, we introduce a unique methodology for terrain map generation, structured in two primary phases. Initially, we employ established procedural map generation techniques, such as Perlin noise and the averaged smoothed sums of Gaussian noise. Subsequently, we implement a Neural Style Transfer, applying style attributes from real-world height maps onto the algorithmically generated image. This process marries the efficiency of procedural generation with the nuanced realism inherent to neural methods. Our findings demonstrate the feasibility of transferring terrain morphological information through Neural Style Transfer. Consequently, this method enables the creation of terrain maps with arbitrary content, yet retains the morphological characteristics of a specific real-world region. This offers a promising alternative to traditional techniques in simulating authentic terrain maps.

In this paper, we commence by outlining the foundational principles of procedural map generation and Neural Style Transfer. We then detail our specific map generation techniques and offer an in-depth overview of our implementation. This leads into a comprehensive discussion of our findings, ending with our conclusions.

\section{Background}
\label{sec:Back}
In our work we utilize both traditional procedural generation algorithms and the Neural Style Transfer technique. Our structures of choice for representing the generated terrain are height maps, which are a widely used technique for depicting 3D terrain in computer graphics and geospatial simulations. Each pixel's grayscale value in the height map corresponds to a specific elevation in the terrain, with darker shades indicating lower altitudes and lighter shades representing higher elevations. By mapping these grayscale values to vertical displacement, a three-dimensional surface can be generated from a two-dimensional image. This approach allows for efficient storage and quick processing, making it especially suitable for real-time rendering in video games and simulations. Moreover, height maps provide an intuitive way for artists and developers to sculpt and modify terrains, blending the realms of technical precision and creative expression.

\subsection{Procedural noise maps}
In this section we discuss the techniques we used for generating procedural noise maps in the form of height maps. We discuss two main approaches, explicit and procedural noise making. 
\subsubsection{Explicit noise}
\begin{figure*}
\includegraphics[width=\textwidth]{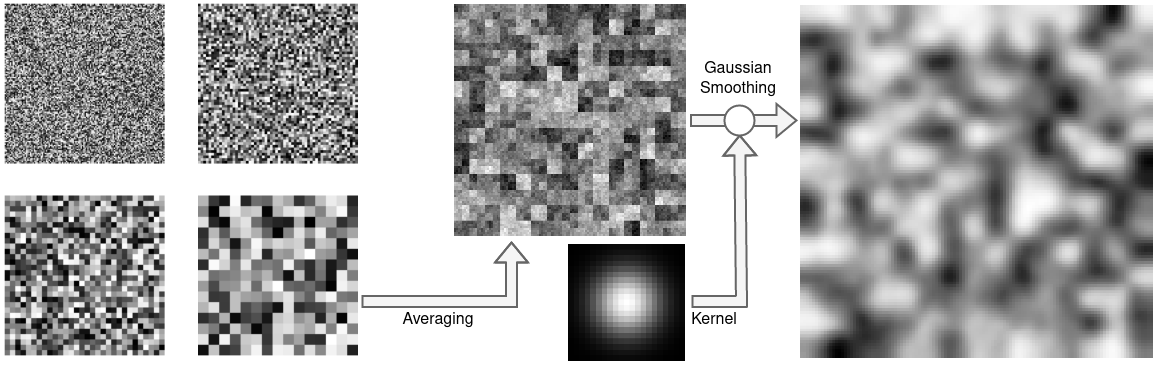}
\caption{Procedural noise generation: averaging multi-scale random matrices, followed by Gaussian kernel smoothing.}
\label{fig:random_generation}
\end{figure*}

By explicit noise we intend the generation of matrices of random and uniformly distributed values \cite{lagae2010survey}. Using these matrices we aggregate noise at different scales such that the noise is not uniformly distributed in the matrix, but creates areas of high and low density.  A convolution operation is then applied to smooth the result, with different kernel and filter sizes at different scales. The process is briefly summarized in Fig. \ref{fig:random_generation}. 
The produced map can be plotted as a 3d surface height map to highlight the terrain structure.
The general procedure for the generation of explicit noise map is is reported in Algorithm \ref{alg:TerrainGeneration}.

\begin{algorithm}
\caption{Explicit noise map generation}
\begin{algorithmic}[1]
\State \textbf{Initialize:} matrices\_list, dimension, sigmas, kernels

\Procedure{ExplicitGeneration}{n}
    \For{i in 1 to n}
        \State matrix = genMatrix()
        \State matrix = upscale(matrix,dimension)
        \For{j in 1 to length(sigma\_list)}
            \State matrix = Convolve(matrix,   sigmas[j], kernels[j])
        \EndFor
        \State matrices\_list.append(matrix)
    \EndFor
    \Return average(matrices\_list)
\EndProcedure

\end{algorithmic}
\label{alg:TerrainGeneration}
\end{algorithm}

\subsubsection{Perlin Noise}
Perlin noise is a well known gradient technique developed by Ken Perling \cite{perlin2002improving}. In our implementation, we leverage this algorithm to produce patterns that exhibit a more organic appearance compared to explicit noise. This approach introduces a dynamic range of variability in the generation process, yielding a diverse array of results matching various circumstances and requirements. The general structure of the perlin algorithm is briefly summarized in Algorithm \ref{alg:PerlinNoise}.

\begin{algorithm}
\caption{Perlin Noise map Generation}
\label{alg:PerlinNoise}
\begin{algorithmic}[1]
\State \textbf{Initialize:} gradient\_list, freq, amplitude

\Procedure{PerlinGeneration}{x, y}
    \State total = 0
    \For{each grad in gradient\_list}
        \State total += Interpolate(grad, x * freq, y * freq) * amplitude
        \State freq *= 2
        \State amplitude *= 0.5
    \EndFor
    \Return total
\EndProcedure

\end{algorithmic}
\end{algorithm}

\begin{figure}
\centering
\includegraphics[width=\linewidth]{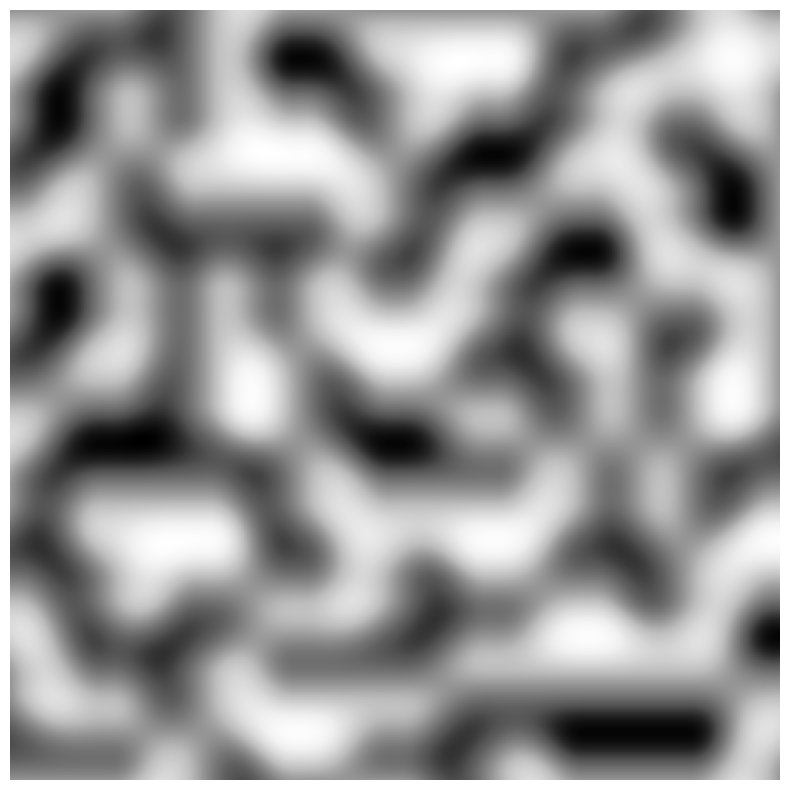}
\caption{Example of a generated Perlin map}
\label{fig:perlin}
\end{figure}

The produced results proved to be relevant to our main task of obtaining a content for the transfer of terrain morphology. Our focus is on developing a texture capable of assimilating style, and Perlin Noise contributes elements of regularity that are advantageous for this aim. An example of a Perlin pattern with a color map applied is reported in Fig. \ref{fig:perlin}.

\subsection{Neural Style Transfer}
Neural Style Transfer is a technique that applies the artistic style of one image to the content of another using a pre-trained convoluted neural network. In our implementation we utilized the VGG-19  \cite{simonyan2014very}architecture.

The definition of content relies on the ability of the network to classify images, and it is defined as the output of one of the last convoluted layers of the network, for each of the two different images.

Style relies on a more complex abstraction, and it is defined as the correlation between the different filter responses at multiple layers of the convoluted network. 
The correlations are computed with the use of Gram matrices.
Practically, we define $F^l \in R^{N_l x M_l}$ as the response of a layer $l$ where $F^l_{ij}$ is the activation of the $i^{th}$ filter at position $j$ in layer $l$, where a layer has $N_l$ feature maps each of size $M_l$.

For the definition of content we rely on the response of the $2^{nd}$ convoluted layer of the $5^{th}$ group, called conv\_5\_2. We therefore define the content loss as the quadratic loss of the features representation of the two images at the given layer, called $F1$ and $F2$
\begin{eqnarray}
L_{content} = \sum_l{(F1^{l}_{ik} - F2^{l}_{jk})^2}
\end{eqnarray}
We use this loss to apply gradient descent on the original content picture, in such a way that the loss is minimized and the content preserved. In our implementation we give a low weight to this loss, considering that the content consistency of the randomly generated model is not as relevant as the correct transfer of the morphological information.

Style is defined as the correlation between the different filter responses of multiple layers of the network, the correlations are given by a gram matrix defined as follows:  
The Gram matrix $G^l \in R^{N_l x N_l}$ where $G^{l}_{ij}$ is the inner product between the vectorised feature map $i$ and $j$ in layer $l$:
\begin{eqnarray}
G^{l}_{ij} = \sum_l{F^{l}_{ik} F^{l}_{jk}}
\end{eqnarray}
We define the style loss as the quadratic loss of the two gram matrices obtained with the original and the generated images, called respectively $G1$ and $G2$. 
\begin{eqnarray}
L_{style} = \frac{1}{4 N^2_l N^2_l}\sum_{ij}{(G1^{l}_{ij} - G2^{l}_{jj})^2}
\end{eqnarray}
The chosen layer on which to compute the style are commonly the first convoluted layer for each group, namely conv\_1\_1, conv\_2\_1, conv\_3\_1, conv\_4\_1, conv\_5\_1. The final optimization loss is then obtained by the sum of style and content components, utilizing two parameters $\alpha$ and $\beta$ to regulate the balance between the two components and given the random map $r$, the terrain map $t$, and the output image $o$.

\begin{eqnarray}
Loss = \alpha L_{content}(r,o) + \beta L_{style}(t,o) 
\end{eqnarray}

\section{Methods}

\begin{figure}
    \centering
    \begin{subfigure}{0.9\linewidth}
        \includegraphics[width=0.95\linewidth]{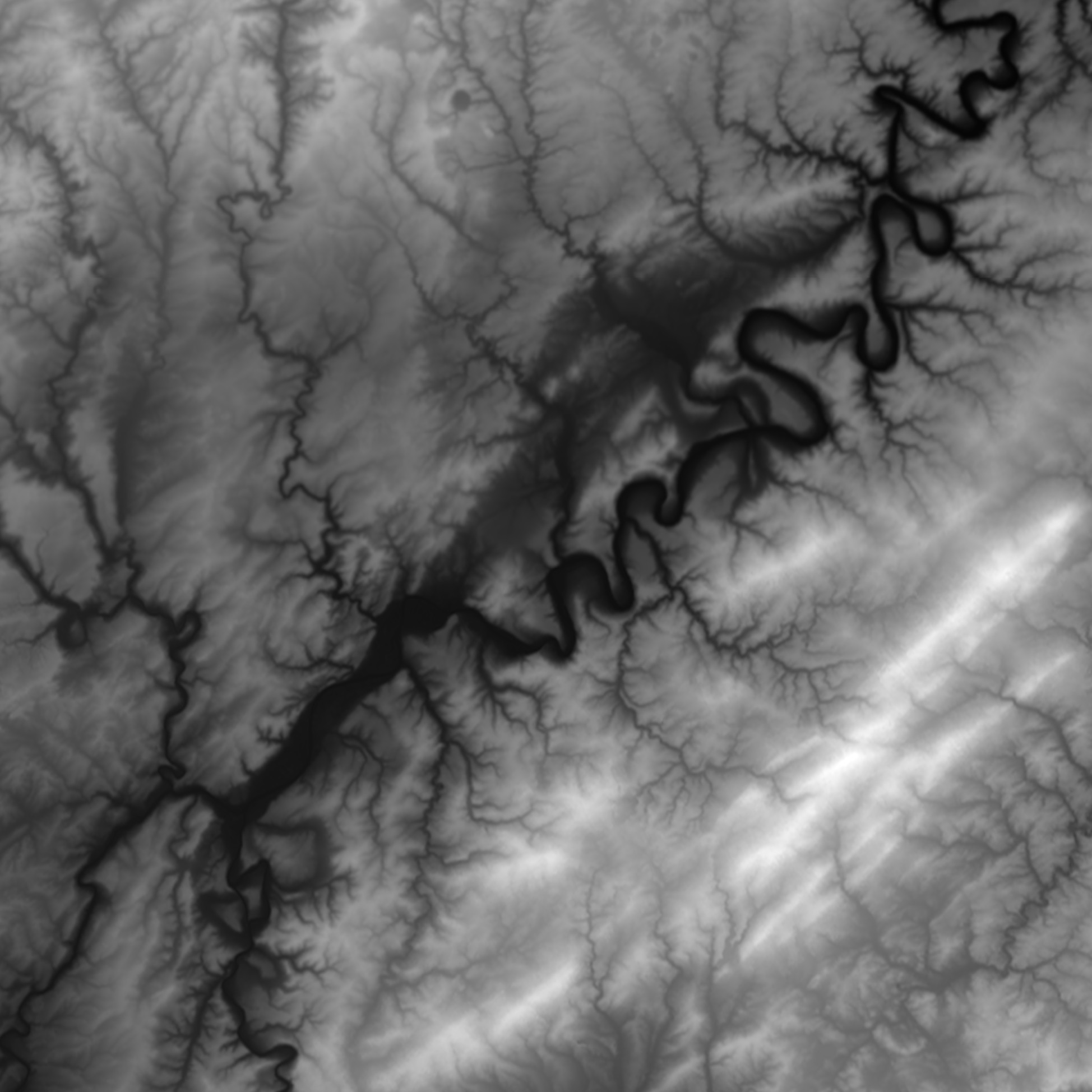}
        \caption{source height map with river morphology}
        \label{fig:himalaya1}
    \end{subfigure}
    \begin{subfigure}{0.9\linewidth}
        \includegraphics[width=0.95\linewidth]{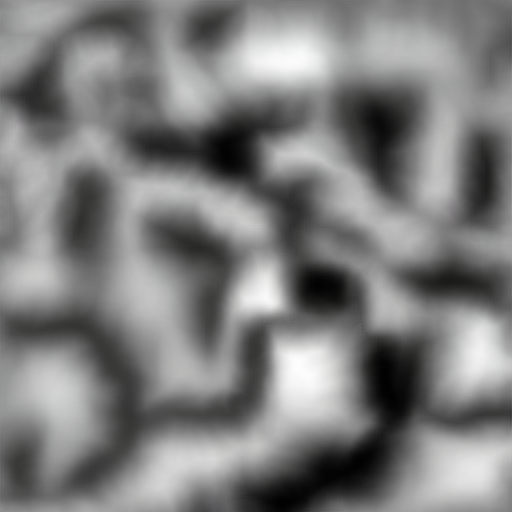}
        \caption{procedural generated noise map}
        \label{fig:himalaya2}
    \end{subfigure}
    \begin{subfigure}{0.9\linewidth}
        \includegraphics[width=0.95\linewidth]{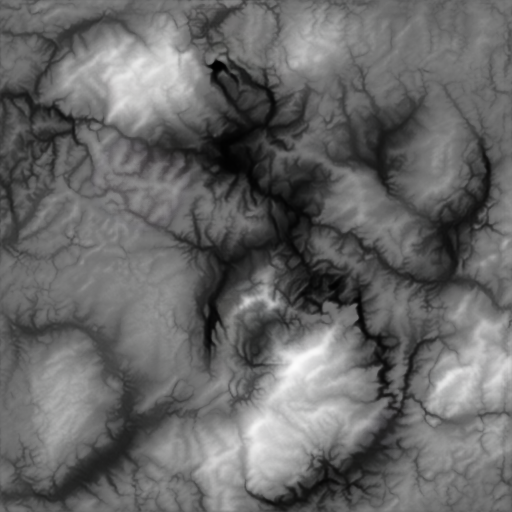}
         \caption{map with transferred morphology}
         \label{fig:generation_example_content2}
        \end{subfigure}
    \caption{Morphological transfer of river-like morphology, starting from the real-world source (a) a procedural noise map (b) and the synthesized custom image (c)}
    \label{fig:river_result}
    
\end{figure}

In this section, we outline our methodology for terrain map generation. Our approach consists in combining a procedural generated image with style elements obtained from real-world height maps, by transferring morphological properties with the use of neural style transfer. For the generation of the procedural maps we apply either one of two methods, namely explicit noise and Perlin noise described in Section \ref{sec:Back}. Both methods provide effective content on which transfer morphological traits: the first exhibits a more random pattern, while the second offers a distinctly organic appearance.
For the task of transferring terrain morphological information, we need to find a terrain map from which to extract the terrain information. Ensuring map uniformity is crucial, given that the gram matrices encoding style are non-spatial in nature. 

In our implementation we use the VGG-19 pre-trained convoluted neural network \cite{simonyan2014very}, excluding the two final dense layers, which provides us a feature space made of 16 convoluted layers. The network is set up with 5 groups of convoluted layers separated by 4 max pooling layers. The first two groups contain 2 convoluted layers each and the latter 5 layers each, denoted \textit{conv\_a\_b} where $a$ represents the group and $b$ represents the layer.

Having noted local inconsistencies, we add a total variation (TV) term in the Neural Style Transfer loss, which help keeping the generated image locally coherent. This loss is computed directly on pixel coordinates, making values of neighboring pixels on the synthesized image closer.  
Given pixel value $x_{ij}$ at the coordinates $(i,j)$ we define the total variation loss as: 
\begin{eqnarray}
L_{TV} = \sum_{ij}{  \vert x_{i,j} \vert }- x_{i+1,j} +  \vert x_{i,j} - x_{i,j+1} \vert
\end{eqnarray}

We can therefore define the final loss, given the random map $r$, the terrain map $t$, and the output image $o$:
\begin{eqnarray}
Loss = \alpha L_{cont.}(r,o) + \beta L_{style}(t,o) + \gamma L_{t.v.}(o)
\end{eqnarray}

With $\alpha,\beta,\gamma$ as the tunable weights of respectively content, style and total variation. Which in our case were set to the following values, increasing the style weight at the expense of content:
\begin{eqnarray}
\alpha = 1 \times 10^-5, 
\beta = 2.5 \times 10^-11,
\gamma = 1 \times 10^-10
\end{eqnarray}

For the transfer of morphological features we need to obtain an image from which the terrain morphology can be pulled. In our work we relied on terrain height maps obtained from the publicly available on the open source project Tangrams Heightmapper \cite{tangrams}.
Our general procedure for the creation of a terrain map can thus be synthesized in the following procedure:  
\begin{enumerate}
    \item Obtain a height map from the real world suitable for morphological transfer.
    \item Create a procedural noise maps able to receive the extracted information, applying the procedures described above.
    \item Tune and execute style transfer algorithm using the procedural noise as content and the height map from the real world as style. 
\end{enumerate}  

In the process of making procedural terrain maps we may want to include arbitrary features into the map, such as in artificial terrain that must adhere to some properties for aesthetic or semantic purposes. For this reason a technique for making procedural maps is highly regarded if enables the final user to provide inputs steering the procedural generation. In our technique we successfully experimented with a procedure to include this feature, our method consists in averaging a custom drawn map with a procedural generated map, and using this combined map as a source for the style transfer. An figure depicting this process is reported in Figure \ref{fig:custom_map}. 

\section{Results}

\begin{figure}
    \centering
    \begin{subfigure}{0.9\linewidth}
        \includegraphics[width=0.95\linewidth]{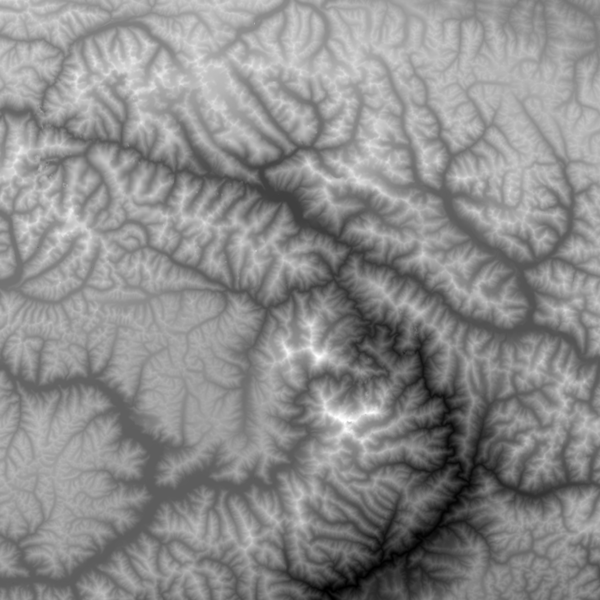}
        \caption{source map with mountain morphology}
    \end{subfigure}
    \begin{subfigure}{0.9\linewidth}
        \includegraphics[width=0.95\linewidth]{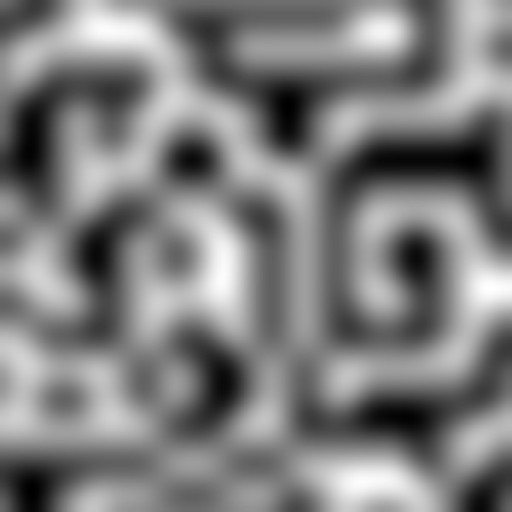}
         \caption{procedural generated noise map}
         \label{fig:generation_example_content3}
        \end{subfigure}
    \begin{subfigure}{0.9\linewidth}
        \includegraphics[width=0.95\linewidth]{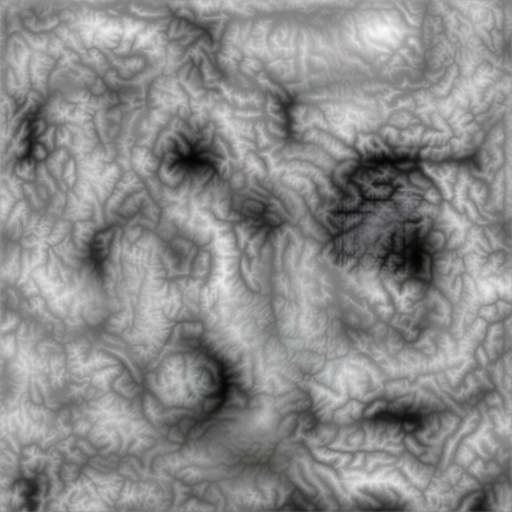}
        \caption{map with transferred morphology}
    \end{subfigure}
    \caption{Morphological transfer of mountain morphology, starting from the real-world source (a) a procedural noise map (b) and the synthesized custom image (c)}
    \label{fig:mountain_result}
    
\end{figure}

In this section, we delve into our experimental setup, analyze the results we achieved, and offer an assessment of the obtained outcomes.

For the utilized neural networks we leveraged pre-trained weights from ImageNet \cite{russakovsky2015imagenet}. As noted earlier, gradient descent is applied directly to the image using SGD. The SGD optimizer incorporates Exponential Decay, and the training process spans 2000 iterations.

For our results we selected three distinct terrain types for our study: mountainous, river, and desert. The height maps for these terrains were sourced from real-world measurements and have a resolution of 1500x1500 pixels. 

\begin{figure}
    \centering
    \begin{subfigure}{0.9\linewidth}
        \includegraphics[width=0.95\linewidth]{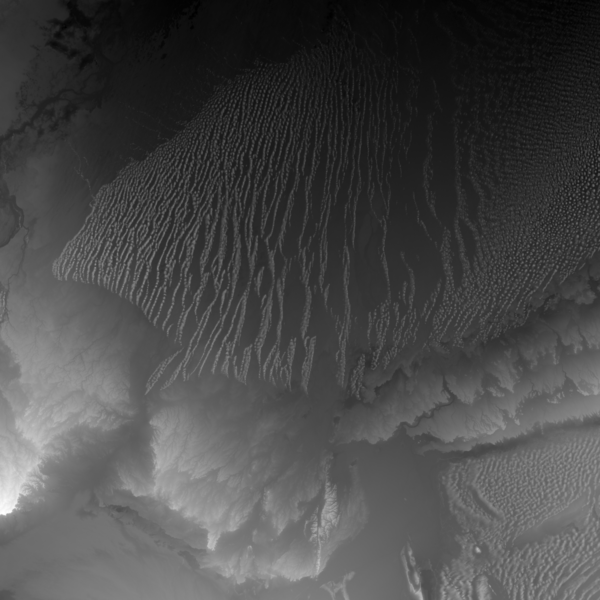}
        \caption{source height map with desert morphology}
    \end{subfigure}
    \begin{subfigure}{0.9\linewidth}
        \includegraphics[width=0.95\linewidth]{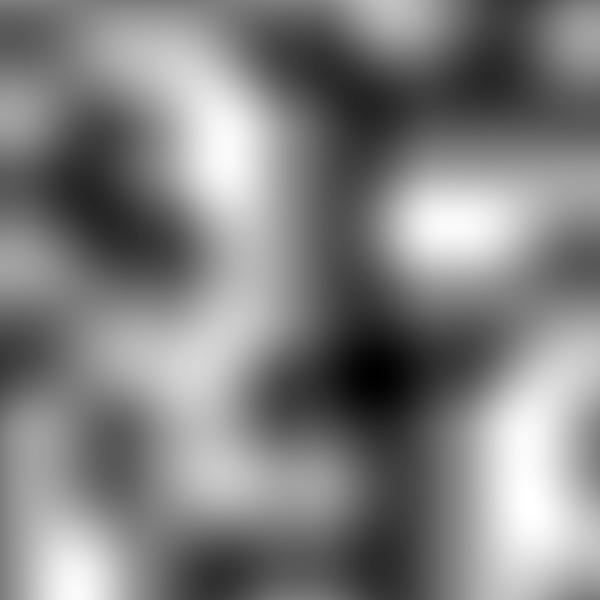}
         \caption{procedural generated noise map}
         \label{fig:generation_example_content1}
        \end{subfigure}
    \begin{subfigure}{0.9\linewidth}
        \includegraphics[width=0.95\linewidth]{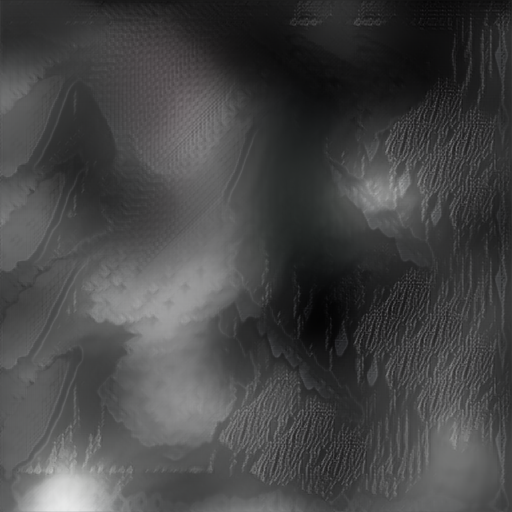}
        \caption{map with transferred morphology}
    \end{subfigure}
    \caption{Morphological transfer of desert morphology, starting from the real-world source (a) a procedural noise map (b) and the synthesized custom image (c)}
    \label{fig:desert_result}
    
\end{figure}

\begin{figure}
    \centering
    \begin{subfigure}{\linewidth}
        \includegraphics[width=\textwidth]{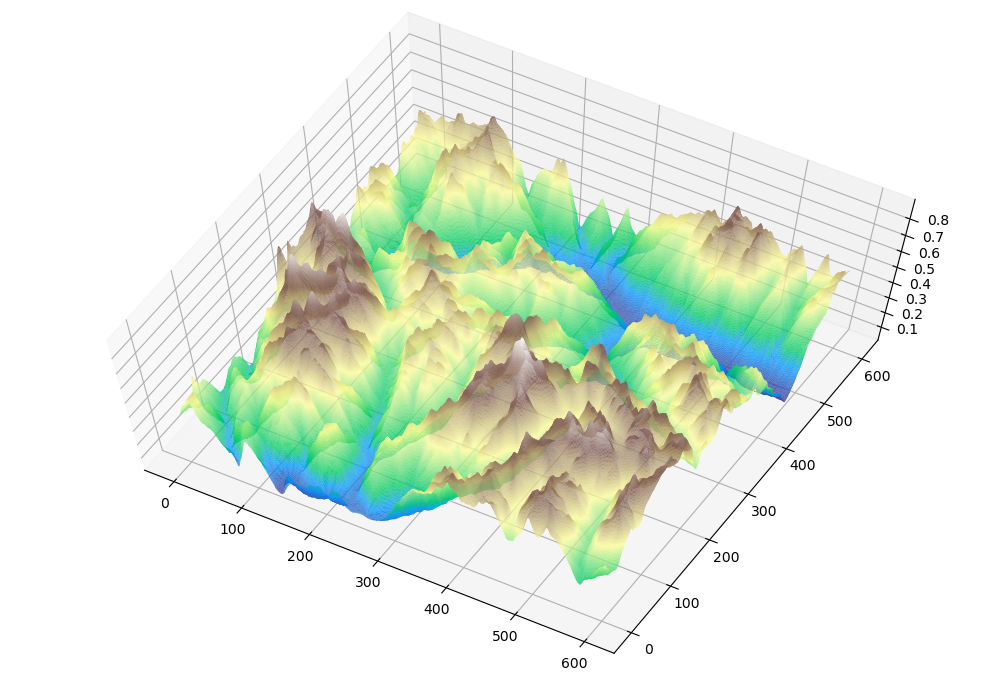}
        \caption{source height map}
        \label{fig:italian_alps}
    \end{subfigure}
    \begin{subfigure}{\linewidth}
        \includegraphics[width=\textwidth]{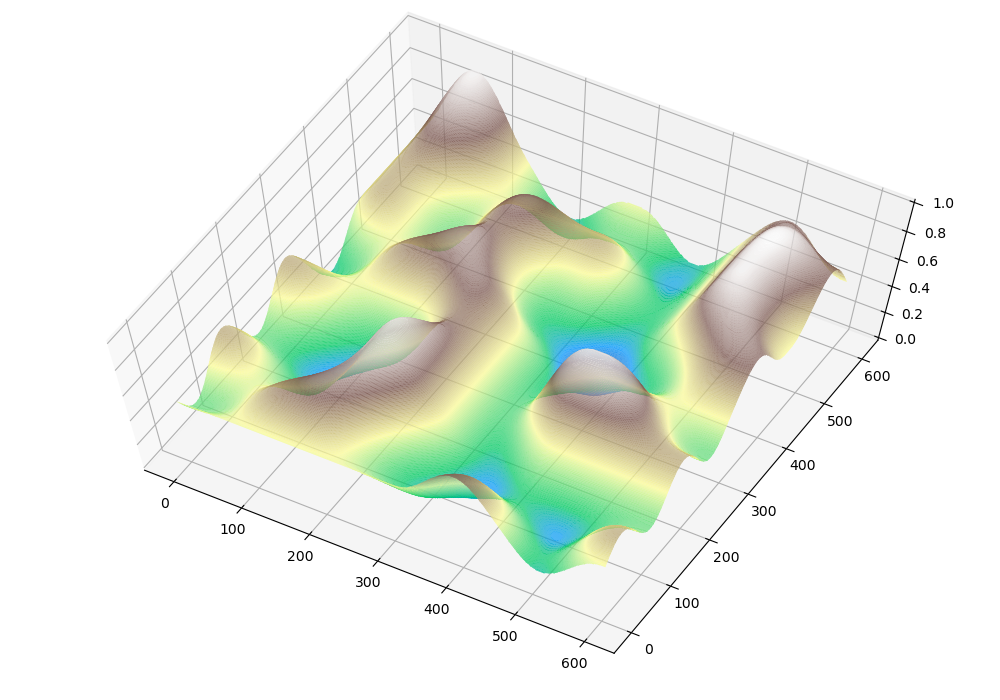}
         \caption{procedural generated noise map}
        \end{subfigure}
    \begin{subfigure}{\linewidth}
        \includegraphics[width=\textwidth]{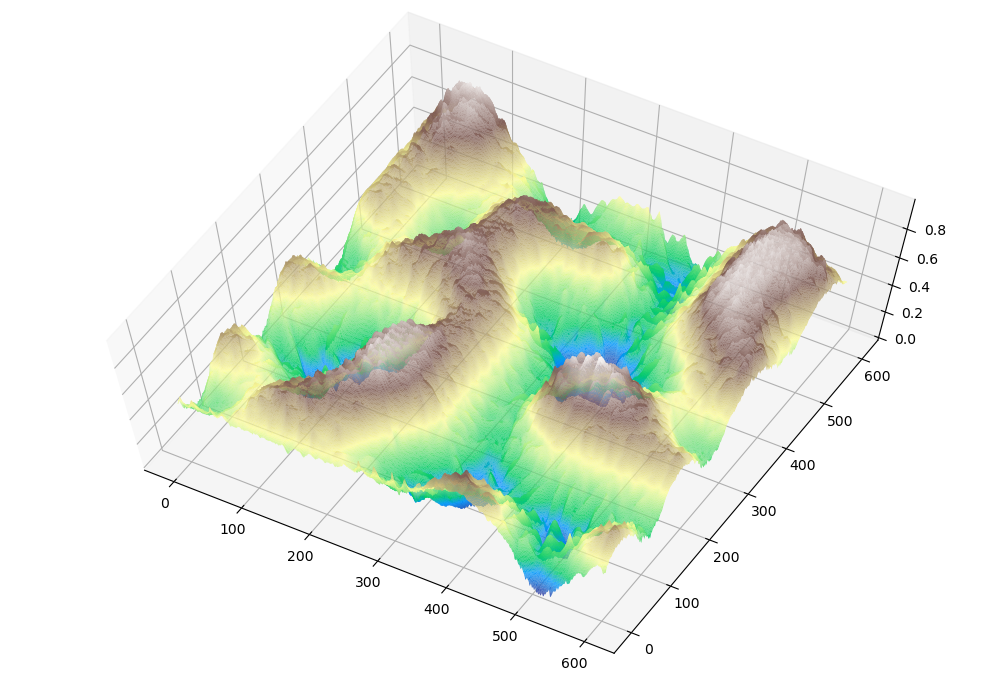}
        \caption{map with transferred morphology}
    \end{subfigure}
    \caption{3D rendering of the source height map (a), the procedural generated noise map (b) and the morphological transferred image (c)}
    \label{fig:3d_generation_example}
\end{figure}

In Figure \ref{fig:river_result}, we showcase an example utilising river morphology. The noise map is crafted through explicit noise. By visually analyzing the map with transferred morphology we note  that river structures have been seamlessly integrated throughout the image, aligning well with the noise map's content distribution. Meanwhile, in Figure \ref{fig:mountain_result}, we demonstrate the transfer of mountainous attributes, drawing from a Perlin-generated noise map and a height map sourced from the Himalayas. The Perlin noise map displays smooth gradients, in contrast to the real-world map, which reflects terrain morphologies shaped by weathering processes. Notably, we successfully imprinted these features onto the generated image, capturing valleys, peaks, and other terrain characteristics, all while preserving the general structure set by the noise map. In Figure \ref{fig:3d_generation_example}, we also present a 3D plot of a sample generation to enhance the visualization of its characteristics. In Figure \ref{fig:desert_result}{} we also report a synthesis utilizing a desert-sourced height map over a explicitly generated noise map, showing the peculiar structures of desert terrain being successfully transferred. 

\begin{figure*}[h!]
    \centering
    \begin{subfigure}{0.3\textwidth}
        \includegraphics[width=\linewidth]{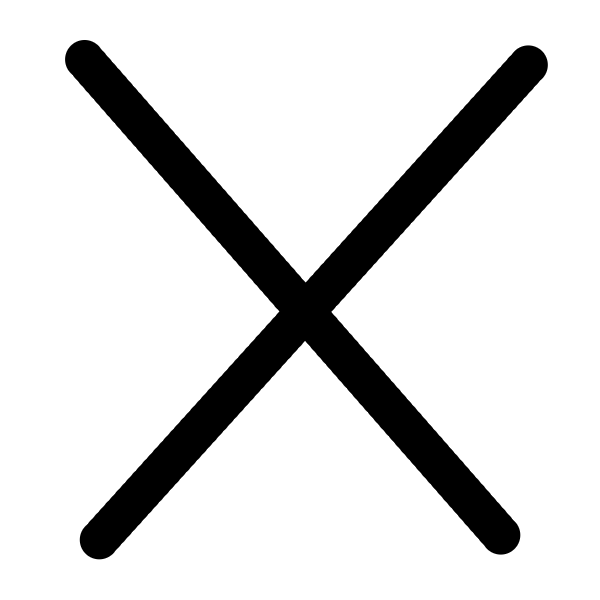}
        \caption{source custom feature}
    \end{subfigure}
    \hfill 
    \begin{subfigure}{0.3\textwidth}
        \includegraphics[width=\linewidth]{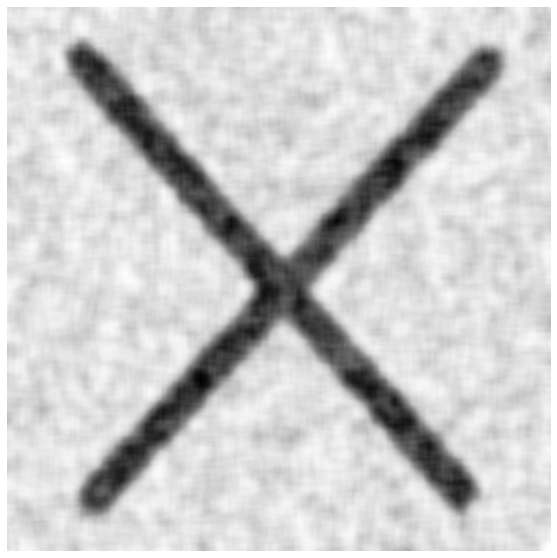}
        \caption{feature with added noise layers}
    \end{subfigure}
    \hfill 
    \begin{subfigure}{0.3\textwidth}
        \includegraphics[width=\linewidth]{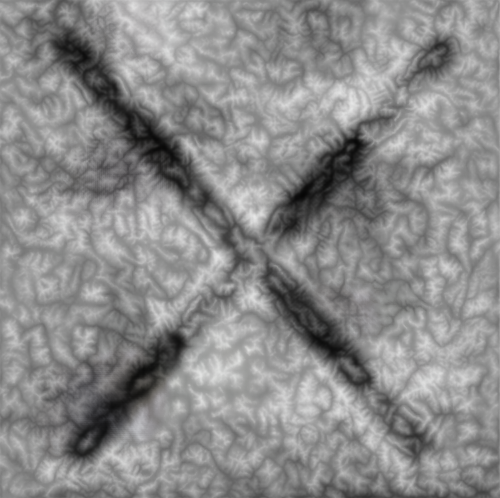}
        \caption{map after morphology transfer}
    \end{subfigure}
    \caption{In this image we report the steps for creating a map with custom features and transferred mountainous morphology. A custom binary map is drawn by the creator (a) and subsequently injected with layers of explicit noise (b). We then apply the Neural Style Transfer, selecting an appropriate source from real world maps, and thus generating the final terrain (c). }
    \label{fig:custom_map}
\end{figure*}

A relevant feature of our approach is the possibility of the generation to be conditioned by custom made image content. In the same way that a random noise map condition the general structure of the generated image, a custom drawn map can be integrated into the generation process to create maps having specific semantic features. An example of this approach is reported in Figure \ref{fig:custom_map}, where a drawn element of a cross is integrated into the explicit noise generation steps to create a final image containing mountainous morphological features including the cross. 

Evaluating the quality of a generative model numerically is inherently challenging. In pursuit of a viable assessment method, we utilized the Structural Similarity (SSIM) metric to gauge the resemblance between the original source image and the generated map, both before and after the morphology transfer process. This approach underscores how closely an image mirrors its original terrain counterpart. The outcomes of this experiment are presented in Table \ref{tab:SSIM}. Notably, the morphologically transferred images demonstrate a greater similarity to the original terrains across all three tested terrain types.

\begin{table}[htbp]
\centering
\def\arraystretch{1}
\begin{tabular}{p{5cm}c}
\toprule
\multicolumn{2}{c}{\textbf{SSIM Numerical Evaluation}} \\
\midrule
\textbf{Terrain Map} & \textbf{SSIM $\uparrow$} \\ 
\midrule
mountain source   & 1   \\
procedural noise map                 & 1.80e-01 \\ 
map with transferred morphology       & 2.08e-01 \\ 
\midrule
river source   & 1   \\
procedural noise map               & 2.32e-01 \\ 
map with transferred morphology       & 2.41e-01 \\ 
\midrule
desert source  & 1   \\
procedural noise map                  & 3.46e-01 \\ 
map with transferred morphology       & 4.68e-01 \\

\bottomrule
\end{tabular}
\caption{SSIM numerical analysis comparing procedural noise maps before and after neural style transfer with real-world terrain source. }
\label{tab:SSIM}
\end{table}

\subsection{Performance review}
Evaluating the effectiveness of generative methods is challenging due to the nuanced nature of their results. In this section, we explore this topic by contrasting it with the GAN approach, aiming to provide a clearer perspective on the potential of this technique.

In our general procedure the computational load is distributed mostly on the style transfer step, considering that procedural noise algorithm are very efficient. In our setting we run the gradient loop for 2000 iterations on a RTX 3060 GPU, and it completes the process in 2 minutes and 46 seconds. The computational cost is low, but it is relative to every generated image. On the other hand, GAN based methods behave very differently, usually having a very high computational cost in the training phase and a small computational footprint in the generation phase. 

Example in literature reveal that the training part of a terrain generation GAN model, for a specific input region, can be completed in 16 hours with a cluster of 8 GTX 1080 TI, while the generation step is subsequently is completed in 50ms on a single GTX 1080 TI \cite{panagiotou2020procedural}. 

In conclusion, the GAN model can be very effective in producing terrain maps quickly, however the computational cost for the training on a new region is prohibitive. On the other hand, our method is more accessible, requiring a much lower computational cost for the creation of an image, but scales poorly when many images are needed. 

\section{Conclusion}
In this paper we demonstrated the feasibility of style transfer applied to morphological terrain information. We utilize this knowledge to successfully synthesize terrain maps with the help of procedural terrain generation algorithms. 

The application of this technique can be exploited for the creation of virtual environments and territories, by pulling the morphological features from a map of the real world. The obtained results showcase the capability of successfully generating terrains with a variety of morphological characteristics, and numerical assessments confirm the the generated images have a higher fidelity to the original real-world maps with respect of purely procedural generated sources.

Our innovative approach holds promise for various fields. In the realm of video game development, it offers a novel means of generating expansive, detail-rich worlds without the exhaustive need for manual design, providing developers with access to custom, morphologically coherent maps that are both simple and efficient to create. Additionally, architectural visualization and urban planning could greatly benefit from this technique.

The potential applications of our method extend to film and animation, geographical information systems, virtual reality, and scientific research. In the context of climate change, this technique could be instrumental in estimating the effects of atmospheric agents on specific territories such ass weather erosion, especially when complemented with additional territorial data impacted by these agents.

Our technique stands out for its ability to create custom maps with relatively low computational demand, making this technology accessible to a broad audience and suitable for various creative purposes. While our method presents a higher computational cost per image compared to other generative terrain generation approaches, it eliminates the typically prohibitive training costs. Although the per-image cost is a limitation, we assert that our tool is a viable solution for small-scale applications and is more suited to individual users than large-scale applications.

In conclusion, our research highlights a novel and versatile approach to terrain generation. Future work could explore the integration of style-conditioned Diffusion Models to further streamline the generation process into a single, efficient step.

\subsection*{Acknowledgements}
This research was partially funded and supported by Future AI Research (FAIR) project of the National Recovery and Resilience Plan (NRRP), Mission 4 Component 2 Investment 1.3 funded from the European Union - NextGenerationEU.

\subsection*{Statements and Declarations}
The author declare no competing interests.

\subsection*{Code availability}
The code relative to the presented work is archived in the following GitHub \href{repository}{https://github.com/fmerizzi/Procedural-terrain-generation-with-style-transfer}.


\bibliography{bib}

\end{document}